\documentclass[pdflatex,sn-apa,iicol]{sn-jnl} 
\usepackage[T1]{fontenc}
\usepackage{lmodern}     
\usepackage{fix-cm}      
\usepackage{graphicx}
\usepackage{multirow}
\usepackage{amsmath,amssymb,amsfonts}
\usepackage{amsthm}

\usepackage[title]{appendix}
\usepackage{xcolor}
\usepackage{textcomp}
\usepackage{manyfoot}
\usepackage{booktabs}
\usepackage{algorithm}
\usepackage{algorithmicx}
\usepackage{algpseudocode}
\usepackage{listings}
\usepackage{enumitem}
\usepackage{makecell}
\usepackage{float}

\begin{document}
\title[Article Title]{On the Generalizability of Iterative Patch Selection for Memory-Efficient High-Resolution Image Classification}

\author*[1]{\fnm{Max} \sur{Riffi-Aslett} \href{https://orcid.org/0009-0004-2683-5652}{\textsuperscript{ORCID}}}\email{max.riffi3@gmail.com}
\author[1]{\fnm{Christina} \sur{Fell} \href{https://orcid.org/0000-0003-0590-4132}{\textsuperscript{ORCID}}}\email{cmf21@st-andrews.ac.uk}

\affil*[1]{\orgdiv{School of Mathematics and Statistics}, \orgname{University of St Andrews}, \orgaddress{\postcode{KY16 9SS}, \country{Scotland}}}

\abstract{Classifying large images with small or tiny regions of interest (ROI) is challenging due to computational and memory constraints. Weakly supervised memory-efficient patch selectors have achieved results comparable with strongly supervised methods. However, low signal-to-noise ratios and low entropy attention still cause overfitting. We explore these issues using a novel testbed on a memory-efficient cross-attention transformer with Iterative Patch Selection (IPS) as the patch selection module. Our testbed extends the megapixel MNIST benchmark to four smaller O2I (object-to-image) ratios ranging from 0.01\% to 0.14\% while keeping the canvas size fixed and introducing a noise generation component based on Bézier curves. Experimental results generalize the observations made on CNNs to IPS whereby the O2I threshold below which the classifier fails to generalize is affected by the training dataset size. We further observe that the magnitude of this interaction differs for each task of the Megapixel MNIST.  For tasks "Maj" and "Top", the rate is at its highest, followed by tasks "Max" and "Multi" where in the latter, this rate is almost at 0.  Moreover, results show that in a low data setting, tuning the patch size to be smaller relative to the ROI improves generalization, resulting in an improvement  of + 15\% for the megapixel MNIST and + 5\% for the Swedish traffic signs dataset compared to the original object-to-patch ratios in IPS. Further outcomes indicate that the similarity between the thickness of the noise component and the digits in the megapixel MNIST gradually causes IPS to fail to generalize, contributing to previous suspicions. Our code is made available at \url{https://github.com/MRiffiAslett/ips_MaxRiffiAslett.git}.}

\keywords{High-resolution image classification, deep learning, patch selection, overfitting}

\maketitle

\section{Introduction} 
\begin{figure}[htbp]
    \centering
    \includegraphics[width=\columnwidth]{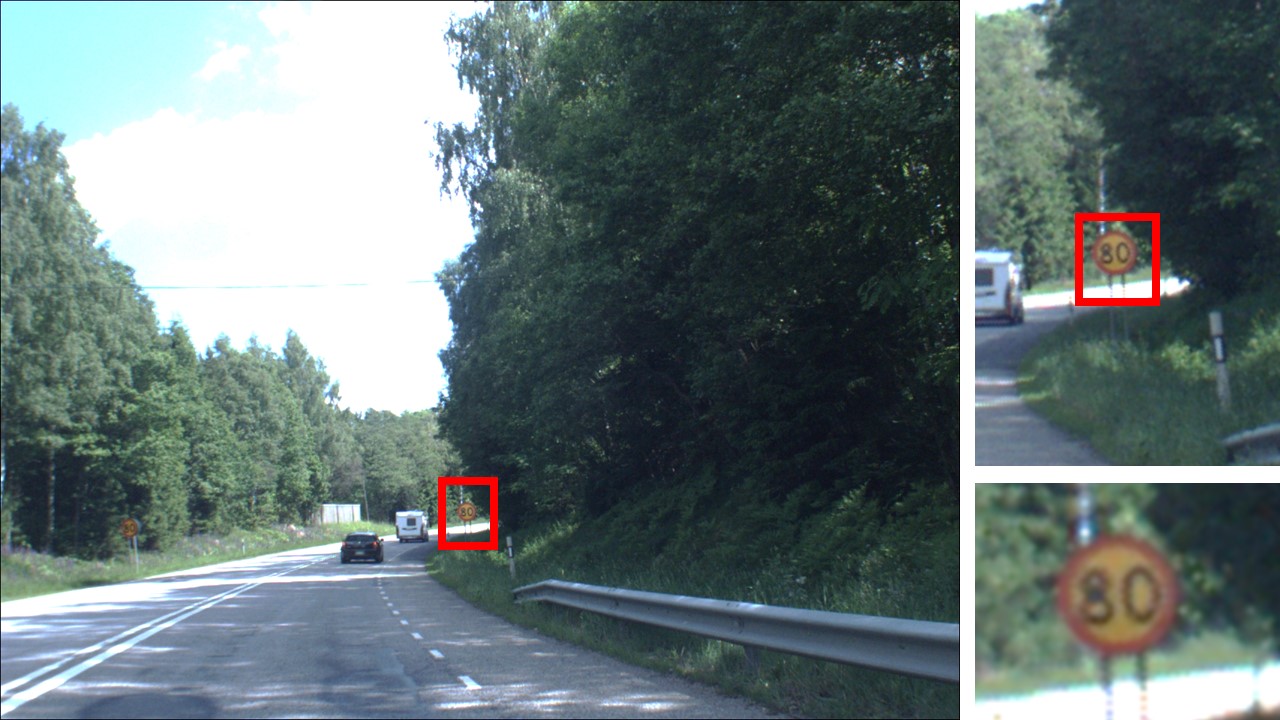}
    \caption{Image ``000036'' from the Swedish traffic signs dataset with label ``80''. On the left, the original image in full resolution (\(960 \times 1280\)). On the right, the image is truncated to approximately \(75 \times 45\) (top) and \(35 \times 15\) (bottom).}
    \label{swedishousemafia}
\end{figure}

Advancements in Convolutional Neural Networks (CNN) such as AlexNet \citep{krizhevsky_imagenet_2012} and ResNet \citep{he_deep_2015} have been successful in classifying natural images with resolutions below one megapixel on datasets such as ImageNet \citep{deng_imagenet_2009}. Yet, in multiple practical applications such as aerial imagery \citep{ofli_combining_2016}, traffic monitoring \citep{lalonde_clusternet_2018}, and automatic industrial inspection \citep{abouelela_automated_2005},  the label correlates with only a small part of the input, leading to a low signal-to-noise ratio (refer to Figure \ref{swedishousemafia} for a visual representation of the Swedish traffic signs dataset \citep{larsson_using_2011}). A common solution is to use strongly supervised learning which  utilizes local region-level annotations from domain experts \citep{dehaene_self-supervision_2020}. In digital pathology however, labels for whole-slide images are frequently captured as part of the diagnostic process. Whereas, region-specific information is not usually generated by pathologists, it would be time intensive and therefore expensive to collect \citep{gadermayr_multiple_2024}. 

Weakly supervised methods offer an alternative solution that can be applied to gigapixel images without fully annotated data, only including slide level labels. These methods, however, process all tissue patches at full resolution, increasing memory usage \citep{katharopoulos_processing_2019}. 

To decrease memory consumption, a disjoint but related line of work that also falls under the weakly supervised paradigm, focuses on the fact that it is unnecessary to process the whole input image as relevant information is often unevenly distributed. \cite{katharopoulos_processing_2019} and \cite{kong_efficient_2021} for instance sample patches based on their attention values at low magnification to be processed at higher resolutions. Alternatively, \cite{cordonnier_differentiable_2021} create a discrete ranking of the most informative patches to select and aggregate the top K most salient.

Iterative Patch Selection (IPS),  \citep{bergner_iterative_2023} processes full resolution patches in batches retaining the top M most salient after each iteration. The most informative patches are then aggregated by a cross-attention-based pooling operator reminiscent of Multiple Instance Learning. IPS has achieved state-of-the-art results on the megapixel MNIST, Swedish traffic signs, and CAMELYON16 dataset while boasting lower memory consumption compared to its predecessors \citep{cordonnier_differentiable_2021}, \citep{katharopoulos_processing_2019}.

While these memory-efficient classifiers have achieved comparable results to weakly supervised state-of-the-art methods, they struggle with generalizability in low signal-to-noise scenarios and tend to over-concentrate on a small subset of informative patches, which is intrinsically linked to overfitting \citep{bergner_iterative_2023, kong_efficient_2021, thandiackal_differentiable_2022}.

We contribute to exploring these limitations through extensive experiments aimed at identifying the shortcomings of a state-of-the-art patch-based memory-efficient classifier.  

Our work investigates the robustness of Iterative Patch Selection (IPS) \citep{bergner_iterative_2023} by introducing novel adverse conditions, including a noise generation component and a novel implementation of the Megapixel MNIST benchmark with a varying object-to-image ratio. These experiments led to a deeper investigation into the role of the Object-to-Patch ratio in IPS. Our key contributions are as follows:

\begin{enumerate}[leftmargin=10pt, rightmargin=10pt]
    \item We extend the Megapixel MNIST to a larger canvas where the object-to-image (O2I) ratio changes by keeping the canvas size fixed. We further replace the linear noise components of the Megapixel MNIST implementation by \citep{bergner_iterative_2023} with non-linear Bézier curves aimed at providing more adverse noise conditions. Our results show that performance worsens in low-data scenarios as the O2I ratio drops and that the task of the Megapixel MNIST affects this relationship.
    \item We study the effects on convergence of the proposed non-linear noise components and find that as the thickness of the lines approaches the thickness of the digits, IPS gradually fails to converge.
    \item We evaluate the effects of patch size on generalization in low-data scenarios with low O2I ratios. Findings indicate that IPS benefits from a patch size smaller than or equal to the region of interest on the Megapixel MNIST and Swedish traffic signs dataset.
\end{enumerate}

\section{Related Works}
\subsection{Weakly Supervised Learning}
A common solution to the weakly supervised problem in high-dimensional image classification is Multiple Instance Learning (MIL). It involves assigning labels to collections of instances, called bags. The unordered instances in this case are square regions of the image called patches that are assigned the same label.  Patches are then processed separately by a feature extractor that usually consists of a CNN. Their outputs are then aggregated by a pooling function consisting of any differentiable function \citep{gadermayr_multiple_2024}. 

More related to our work on IPS,  transformer-based MIL methods were introduced to capture the correlations between different instances. Correlated MIL \citep{shao_transmil_2021}, employs self-attention \citep{vaswani_attention_2023}, to aggregate instance-level features into a bag-level representation for classifying breast cancer WSI. Alternatively, Attention-based Deep MIL \citep{ilse_attention-based_2018} employs a  Gated Attention (GA) \citep{xue_not_2019} mechanism as the weighting scheme for grading breast cancer and colon cancer whole-slide images.

The aforementioned weakly supervised methods are not intended to lower the computational and memory footprint of patch-based classifiers. Recurrent visual attention models improve computational efficiency by only processing some parts of the full image. The research within this framework sequentially processes patches, to select specific regions for further processing. \cite{mnih_recurrent_2014} first used a recurrent neural network to identify regions of interest in high-resolution images using reinforcement learning to train their model as it is non-differentiable. 

\cite{katharopoulos_processing_2019} introduced an attention network to sample important areas in a down-sampled view of the original image to process a portion of the original image. The sampled patches are then aggregated by computing the expectation of the patches over their attention distribution. \cite{kong_efficient_2021} extend \cite{katharopoulos_processing_2019}'s work by splitting the attention-based sampling process into two stages, sampling patches at a lower resolution and then progressing to the higher resolution.

One limitation of down-sampling the image happens when there is no discriminate information at a lower scale. \cite{kong_efficient_2021} for instance assessed their Zoom-In network's performance on the Needle MNIST dataset \citep{pawlowski_needles_2020}, where the task is to detect the presence of the digit "3" (Refer to Section \ref{needdleMnist}).  Their network failed to handle this dataset as down-sampling the image washes out discriminative information. This in turn prevents their approach from finding regions of interest.

\cite{cordonnier_differentiable_2021} also builds upon the work of \cite{katharopoulos_processing_2019} by introducing a differentiable Top-K operator to select relevant patches in the image. In their work, a shallow scorer network that operates on a downsampled image is used to assign a relevance score to each patch used to select the K most relevant. As discrete rankings are not differentiable, they employ the previously introduced perturbed maximum approach from \cite{blondel_fast_2020}, which incorporates Gaussian noise into each rank to make them differentiable. 

\subsection{Iterative Patch Selection}
\label{IPSsection}
Iterative patch selection (IPS)  \citep{bergner_iterative_2023} is the method implemented in our work. IPS iterates through each patch in the image to only maintain the top M most informative in memory. The image is split into patches and fed through IPS in batches, which loads a fixed number of patches \(I\) at a time. At each iteration, patches are encoded first with a CNN and then with a cross-attention module running in no gradient mode to extract attention values. These attention values allow IPS to keep only the top M most informative patches in memory after each iteration. Once the top \(M\) patches of dimension  \(D\) are selected,  \(X^* \in \mathbb{R}^{M \times D}\) they are aggregated by a weighted average as shown in Equation \ref{1}.

\begin{equation}
    z = \sum_{m=1}^M a_m (X^*_m W^v) \tag{1}
    \label{1}
\end{equation}

where \(a_m\) is the attention score for the \(m\)-th patch and \(X^*_m W^v\) is a linear projection of the patch embeddings  \(X^*\) and \(\mathbf{W}^v \in \mathbb{R}^{D \times D_v}\) is a learnable weight matrix, where \(D_v\) is the dimension of the value embeddings. In transformer notation, \(X^*_m W^v\) corresponds to the values (V). Attention score \(a_m\) are derived from a multiple head cross attention layer which follows the original setup of transformers \citep{vaswani_attention_2023}.

Importantly, at each iteration of IPS,  patches are scored by the same multi-head cross-attention module in no gradient mode. The  attention scores are then used to select to top M most salient patches at each iteration before being fed through the same module again, in gradient mode to obtain a bag-level representation as in equation \ref{1}. A key advantage of IPS over Attention Sampling \citep{kong_efficient_2021, katharopoulos_processing_2019} and Differentiable Top-K \citep{cordonnier_differentiable_2021} is that it does not rely on lower resolution views of the original image to detect salient patches. It is therefore advantageous in that it benefits from mostly better efficiency while bypassing down sampling the canvas which can blur discriminate information for very low O2I ratios.

\subsection{Object-to-image ratio}
While Attention Sampling \citep{katharopoulos_processing_2019},  Differentiable Top-K \citep{cordonnier_differentiable_2021}, and Iterative Patch Selection \citep{bergner_iterative_2023} have reached performance levels comparable to fully supervised techniques, difficulties are attributed to scenarios with low object-to-image (O2I) ratios.  \cite{thandiackal_differentiable_2022} found that their Top-K module tends to overlook extremely small metastases, resulting in the misclassification of WSIs due to low attention. Furthermore, \cite{bergner_iterative_2023} found that performance can decline in IPS, if the signal-to-noise ratio decreases, even when using full-resolution images in the patch selection module. This is demonstrated in their experiments with the megapixel MNIST datasets where the image size was scaled from 1k to 10k pixels. Their results showed that IPS performs with high accuracy up to 8k pixels, and begins to decrease in accuracy from 9k pixels. Importantly, these difficulties are attributed to either the patch selector failing to discern disciminative information or failing to assign attention values to very small but informative regions.  

\subsubsection{Needle MNIST}
\label{needdleMnist}
\begin{figure}[htbp]
    \centering
    \includegraphics[width=\columnwidth]{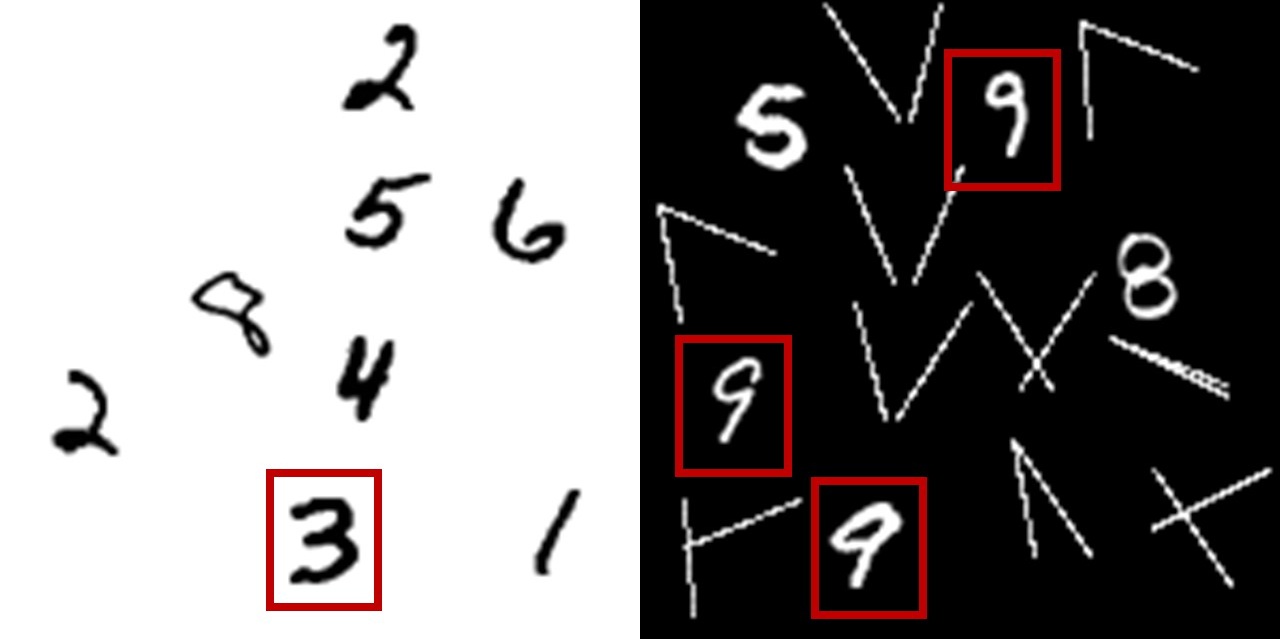}
    \caption{Visual representation of the Needle MNIST dataset (left) and the Megapixel MNIST dataset (right).}
    \label{Tasksvstasks}
\end{figure}

Robustness to small object-to-image ratios was explored previously by \cite{pawlowski_needles_2020} who introduced the Needle MNIST dataset, inspired by the Cluttered MNIST \citep{ba_multiple_2015}. The task is to predict whether the digit "3" appears on the canvas amongst cluttered digits \{0, 1, 2, 4, 5, 6, 7, 8, 9\} sampled with replacement (Refer to Figure \ref{Tasksvstasks}, (left)). 

The object-to-image ratio varies by maintaining the digits at 28 × 28 pixels and increasing the canvas resolution. This results in O2I ratios of \{19.1, 4.8, 1.2, 0.3, and 0.075\}\%, with canvas sizes 64×64, 128×128, 256×256, 512×512, and 1024×1024 pixels. Their findings show that CNNs fail to generalize below a certain signal-to-noise ratio, and the dataset size influences this ratio. 

Our work seeks to follow theirs by reproducing the intuition behind their experiments on IPS, a memory-efficient patch-based classifier. We extend their setup by considering a multi class classification problem where the O2I varies while the canvas size remains fixed. Additionally, our noise generation strategy enables us to generate noise where its resemblance to the region of interest is a parameter that can be controlled.

\subsubsection{Megapixel MNIST}
\begin{figure}[htbp]
    \centering
    \includegraphics[width=\columnwidth]{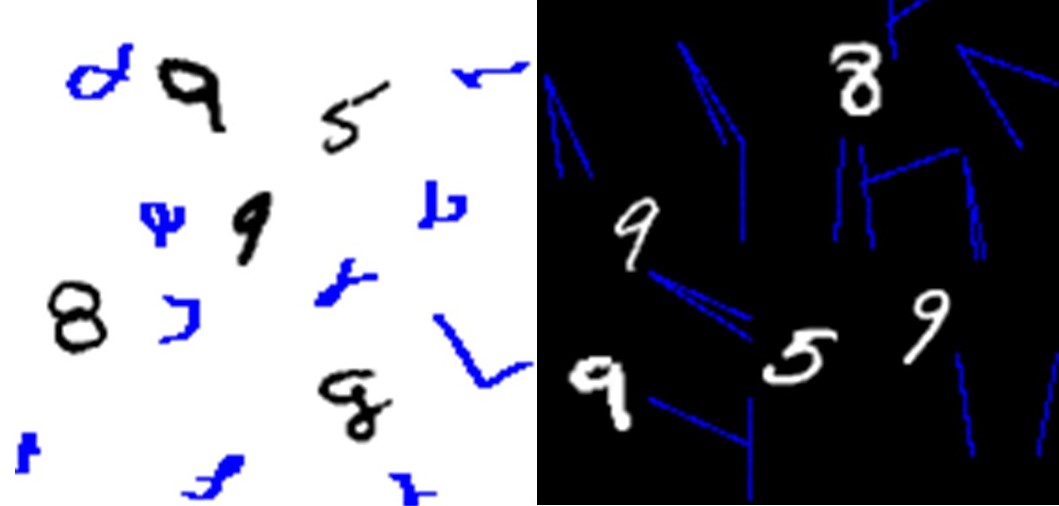}
    \caption{Visual representation of the Megapixel MNIST dataset (150 \(\times\) 150) with 10 noise digits from our method (left) and the original Megapixel MNIST (right).}
    \label{noisegen}
\end{figure}

While \citep{pawlowski_needles_2020}'s Needle MNIST tasks can be solved with just one informative patch, megapixel MNIST necessitates the recognition of multiple patches. The megapixel MNIST dataset introduced by \citep{katharopoulos_processing_2019} features 5 MNIST digits placed randomly on a canvas. Of these, 3 digits belong to the same class, while the remaining 2 are from different classes. The task is to detect the majority class  (refer to Figure \ref{noisegen}, (right)). 

\cite{bergner_iterative_2023} found that this problem is too easily solvable. To introduce more complexity, they extended the setup with three additional tasks: detecting the maximum digit, identifying the topmost digit, and recognizing the presence or absence of all classes.

\section{Noise Generation}
Here we present the noise generation strategy. In the  megapixel MNIST implementation of \cite{katharopoulos_processing_2019}, 50 line patterns are created by sampling angles \(\theta_i\) from a uniform distribution. The slopes \(m_i\) are calculated, and line coordinates \((x_j, y_j)\) are generated based on these slopes, as shown on the right of Figure \ref{noisegen}. The simple lines bear little resemblance to the digits, we instead seek to create nonlinear curves that mimic the structure of digits using Bézier curves (refer to Figure \ref{noisegen}, right). Bézier curves are parametric curves used to model a smooth surface which are defined by the relative positions of a set number of control points \citep{baydas_defining_2019}.  They have been used in computer-aided geometric design for instance to model surfaces with Bézier curves using a shape parameter \citep{qin_novel_2013}.

We suggest randomly sampling control points from a set with predefined probabilities that match the control point counts observed in the digits 1 through 9 (refer to Table \ref{noise_show}).
 
\begin{table}[htbp]
    \centering
    \begin{tabular}{c}
        \includegraphics[width=\columnwidth]{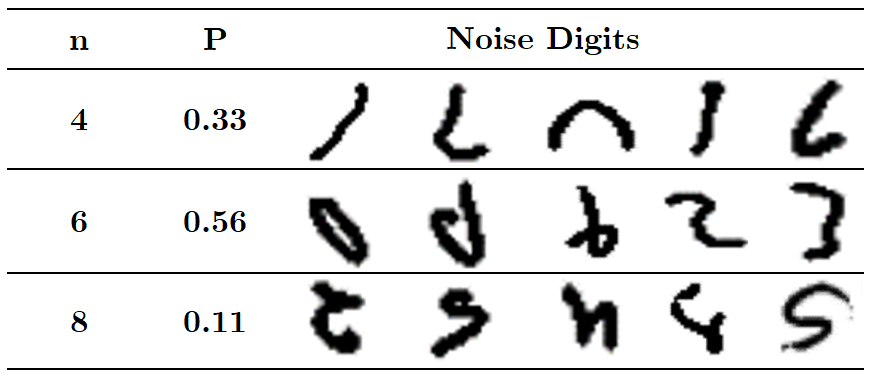} \\
    \end{tabular}
    \caption{Illustration of noise components using Bézier curves with $\{4, 6, 8\}$ control points (n), sampled with probabilities $\left\{\frac{3}{9}, \frac{5}{9}, \frac{1}{9}\right\}$. The two last columns on the right present rare ($<$0.5\% occurrence) noise components chosen for their close resemblance to digits 1, 2, 3, 4, 5, and 6.}
    \label{noise_show}
\end{table}

We begin by sampling control points from a uniform distribution on an $N \times N$ canvas. For each curve, we sample $n$ control points, where $n$ is randomly chosen from the set $\{4, 6, 8\}$ with corresponding probabilities $p$: $\left\{\frac{3}{9}, \frac{5}{9}, \frac{1}{9}\right\}$. We analyzed the number of control points required for each digit to ensure that the sampled number of control points are as close as possible to the control points in digits. For instance, our observations indicated that digit "0" required 4 control points, digit "2" required 6, and digit "8" required 8 control points. 

\begin{algorithm}[htbp]
\begin{algorithmic}[1]
    \State \textbf{Input:} Canvas size \(N \times N\), set of control point probabilities \(\left\{\frac{3}{9}, \frac{5}{9}, \frac{1}{9}\right\}\), number of points \(p \in \{4, 6, 8\}\)
    \State \textbf{Output:} Bézier curves on an \(N \times N\) grid
    
    \State Initialize an empty canvas of size \(N \times N\)
    
    \For{each curve}
        \State Randomly choose \(p\) control points from \(\{4, 6, 8\}\) with probabilities \(\left\{\frac{3}{9}, \frac{5}{9}, \frac{1}{9}\right\}\)
        \State Sample control point positions \(\mathbf{P}_i\) from a uniform distribution on the \(N \times N\) canvas
    
        \State The Bézier curve is defined by the control points:
        \[
        \mathbf{B}(t) = \sum_{i=0}^{n} \binom{n}{i} (1 - t)^{n-i} t^i \mathbf{P}_i, \quad t \in [0, 1]
        \]
        \State Discretize \(t\) into 100 points \(\mathbf{B}(t_j)\) 
    \EndFor
    \caption{Proposed noise generation module using Bézier Curves with a custom sampling scheme to create noise components closer to the distribution of original digits.}
    \label{alg:bezier_noise}
\end{algorithmic}
\end{algorithm}

By aligning the sampled control points with these values, we can generate curved lines with a distribution of control points inspired by what is observed in digits. The Bézier curve as presented by \citep{baydas_defining_2019} is shown in Equation \ref{9}.

\begin{equation}
\mathbf{B}(t) = \sum_{i=0}^{n} \binom{n}{i} (1 - t)^{n-i} t^i \mathbf{P}_i, \quad t \in [0, 1] \tag{9}
\label{9}
\end{equation}

where $\mathbf{P}_i$ are the control points and $n$ is the number of control points minus one. We discretize $t$ into 100 points and draw the  curve on an $N \times N$ grid (refer to Algorithm \ref{alg:bezier_noise}).

\section{Experiments}
For simplicity, our experiments are broken down into 3 questions that our work seeks to answer:

\begin{enumerate}[leftmargin=10pt, rightmargin=10pt]
    \item What observable factors influence the O2I threshold below which IPS fails to generalize, as discussed in Section~\ref{O2Isection}?
    \item Does convergence depend on the resemblance of noise to the Region of Interest (ROI) or the number of noise digits on the canvas? Refer to Section~\ref{digithickness}.
    \item How can adjusting the patch size relative to the ROI (smaller, equal, or larger) be beneficial in limited data scenarios with a low O2I ratio, as explored in Section~\ref{Patchsize}?
\end{enumerate}

For the megapixel MNIST dataset, each model configuration was trained with the following hyperparameters unless specified otherwise: 100 epochs, a batch size of 16, a memory size of 100, an iteration size of 100, a patch size of \(50 \times 50\), patch stride of 50 and a ResNet-18 encoder. 

In experiments with the Swedish traffic signs dataset, each configuration was run for 150 epochs with a memory and iteration size of 10 and 32 patches respectively. The patch size and batch size remain fixed at 100×100 and 16 respectively. Additionally, a ResNet-18 pre-trained on IMAGENET1K V1 in gradient mode is used as the patch encoder. 

For the cross-attention transformer, we followed the hyperparameters of \cite{bergner_iterative_2023}, who in turn used the default values in \cite{vaswani_attention_2023}. The optimization strategy is also unchanged, during the first 10 epochs, the learning rate decreases linearly. When fine-tuning pre-trained networks, the learning rate is adjusted to 0.0003, whereas it is set to 0.001 for networks trained from scratch. Throughout the training process, a cosine schedule is then used to decrease the learning rate by a factor of 1,000 gradually.

\subsection{Object-to-Image Ratio}
\label{O2Isection}
This first part of our work seeks to extend the experiments of \cite{bergner_iterative_2023} in two ways. 1: by varying both the O2I ratio and the size of the training data, unlike \cite{bergner_iterative_2023}, who only changed the O2I ratio, and 2: by changing the noise component of the megapixel MNIST dataset to mimic a more adverse setting. Importantly, our setup keeps the canvas size fixed and scales the digits upward, unlike the approaches in \cite{bergner_iterative_2023} and \cite{pawlowski_needles_2020}, which keep the digit size fixed and scale the canvas size. Their approach operates under the assumption that the number of informative patches relative to the total number of patches is not correlated with validation accuracy. We instead control the canvas size during the experiment offering an alternative methodology. 

We scale the original megapixel MNIST from 1500 by 1500 to 3000 by 3000 pixels and increase the size of the digits. The digit sizes are as follows: 28×28, 56×56, 84×84, and 112×112 pixels, which correspond to the following O2I ratios: \{0.01, 0.034, 0.078, 0.13\}\%. The aforementioned O2I ratios were selected to obtain the largest array of very small object-to-image ratios while fitting within our memory constraints. Additionally, we linearly decrease the noise as a function of the digit sizes as follows: \(\text{noise} = \text{digit size} \times (-7.14) + 1000\), resulting in 800 noise digits for the lowest O2I ratio and 4000 for the highest.

\begin{figure*}[htbp]
    \centering
    \includegraphics[width=\linewidth]{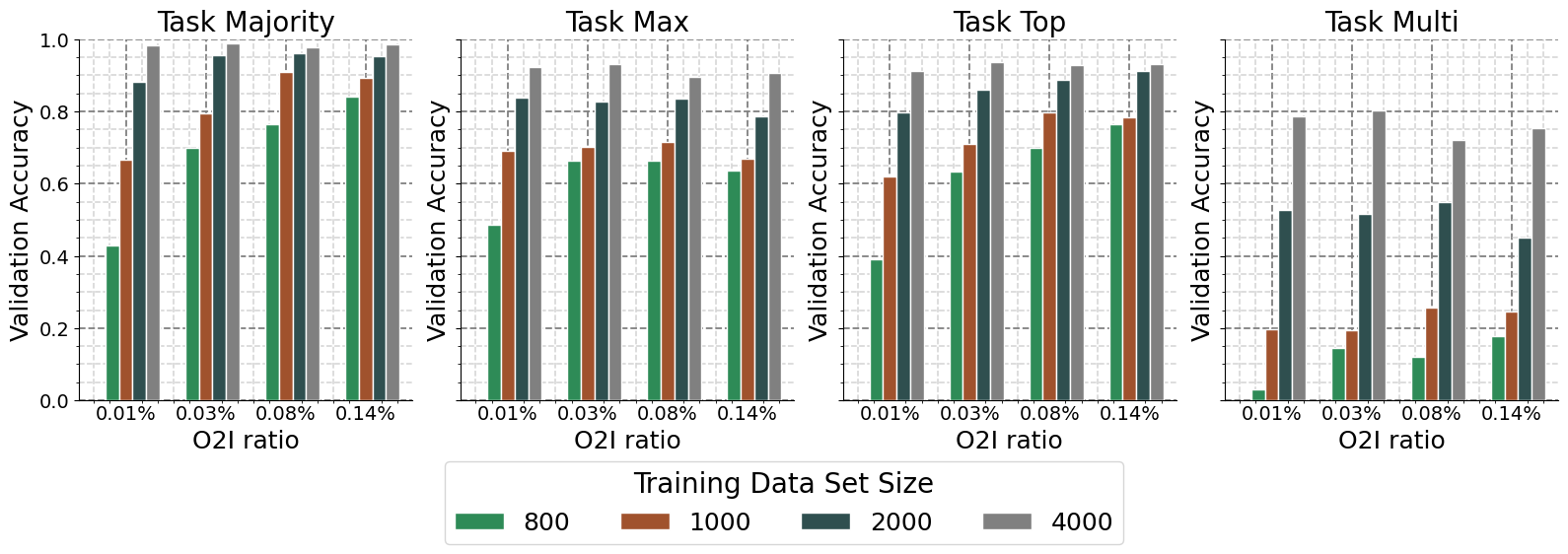}
    \caption{Experiments on megapixel MNIST with a novel noise generation component using Bézier curves that aim to resemble the number of control points found in digits. Four object-to-image ratios were tested: across four training dataset sizes. Canvas size and patch size remain fixed at $3000 \times 3000$ and $50 \times 50$, respectively, and the O2I changes by varying the digit resolutions to $28 \times 28$, $56 \times 56$, $84 \times 84$, and $112 \times 112$.}
    \label{O2I}
\end{figure*}

Each set of O2I ratios is run 4 times with different amounts of training data, specifically \{4000, 2000, 1000, 800\} samples, chosen as 4000 samples solves the task while 800 is the limit below which the model fails to converge.

Changing the object-to-image ratio on the megapixel MNIST dataset (refer to Figure \ref{O2I}) indicates some interactive relationships between the O2I threshold for generalization, the dataset size, and the task of the megapixel MNIST dataset. Notably, for the task of identifying the majority digit ("Maj") and the top-most digit ("Top"), the object-to-image ratio greatly affects the amount of data needed to generalize. For the task Majority for instance, at the lowest O2I ratio (0.01\%), 2000 samples are needed to achieve a validation accuracy of 88\% while at the highest O2I ratio (0.14\%) 1000 samples yield a validation accuracy of 89\%. The same relationship between the dataset size and O2I ratio is also observable for the task "Top".

The tasks "Max" and "Multi" however showcase that the task affects the rate by which the object-to-image ratio affects the number of instances needed to generalize. This rate is much greater for tasks "Maj" and "Top" than it is for tasks "Max" and "Multi".  For the task "Max" for instance, at the lowest O2I ratio (0.01\%), 800 training samples achieve a validation accuracy of  49\% while at the highest O2I ratio (0.14\%), 800 training instances yield a validation accuracy of 64\%. This effect is less present as the training data size increases. With 2000 training instances, for example, the validation accuracy of the task "Max" is 83\% for an O2I ratio of  0.01\% and 79\% for an O2I ratio of 0.14\%.  

For the task "Multi" the rate decreases further. For each number of training instances, no trend is apparent between the O2I ratio and validation accuracy. For instance, at 2000 samples the validation accuracy is $\{79\%, 71\%, 87\%, 83\%\}$ for the following O2I ratios $\{0.01\%, 0.03\%, 0.08\%, 0.14\%\}$.  

Training accuracy for tasks "Maj", "Max", and "Top", converge to 0.99 or higher regardless of the training data size and O2I ratio. The task of identifying each digit, however ("Multi"), only learns the full training data ($>99\%$) when the training dataset size is equal to or greater than 4000. These results suggest that IPS is adept at finding regions of interest in low data scenarios, as shown by the performance of the tasks "Top", "Max" and "Maj", but falls short in identifying each number individually (task: "Multi").   

Our findings indicate that there is a positive correlation between the object-to-image ratio and validation accuracy. The training data size interacts negatively with this effect where the larger the amount of instances used for training, the less generalizability suffers from low object-to-image ratios. By extending the O2I experiments to a multiclass and multitask classification problem, our work finds that the magnitude of the interaction is different for each task of the megapixel MNIST. For tasks "Maj" and "Top" we found that the rate at which the dataset size affects the O2I threshold for generalization was at its highest, followed by tasks "Max" and "Multi" where in the latter, this rate is almost at 0. 

\cite{pawlowski_needles_2020} found that the number of training instances affects the O2I threshold for generalization. We contribute to existing work by finding that this  effect is observable on a patch-based classifier and that there are interactions between the task of the megapixel MNIST and the rate by which the size of the training data affects the O2I threshold for generalization.

\subsection{Effect of noise on convergence}
\label{digithickness}
In \cite{pawlowski_needles_2020}'s experiments, they find that some model configurations fail to converge on both the training and validation sets on the Needle MNIST benchmark, with training accuracy close to random. To further understand this phenomenon, they replace the MNIST digits with Gaussian noise, following the setup in \cite{zhang_understanding_2017}. It was found that although their setup could memorize the Gaussian noise, it failed to recognize the digit "3" from other digits on the canvas, and convergence became increasingly difficult as the object-to-image ratio decreased. They hypothesize that structured noise, such as digits, may be harder for CNNs to process than Gaussian isotropic noise.

\begin{figure}[htbp]
    \centering
    \includegraphics[width=0.8\linewidth]{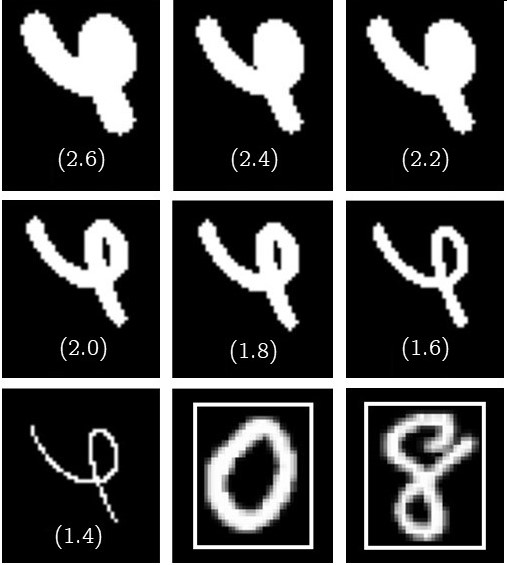} 
    \caption{Visualization of 7 noise digits with the thickness in parenthesis as well as two MNIST digits (bottom right).}
    \label{MNISTbaby}
\end{figure}

Contributing to these observations, we find that as the thickness of the noise digits approaches the thickness of the digits in the megapixel MNIST dataset, IPS fails to converge on both validation and training sets.  We empirically demonstrate this by incrementally changing the thickness of the noise digits to  \(\{1.4, 1.6, 1.8, 2, 2.2, 2.4\}\) (refer to Figure \ref{MNISTbaby} for a visual comparison between the thickness of original MNIST digits and the thicknesses of a noise digit). Each setup was run for 50 epochs across 3 random seeds on a $1500 \times 1500$ canvas with a $28 \times 28$ digit size (O2I ratio: 0.03\%) and noise size of 800. 

Table \ref{digitthick} illustrates that the validation accuracy for all tasks decreases in a non-linear fashion from thickness 1.4 to 1.8 before reaching random accuracy at digit thickness 2. Tasks "Maj" and "Top" do not converge with a random accuracy of 10\% as 10 digits could be chosen. Task multi does not converge at 0\% as there are 10 possibilities for each of the 5 digits which is $10^5 = 100,000$. 

\begin{table}[htbp]
\centering
\setlength{\tabcolsep}{2pt}  
\renewcommand{\arraystretch}{1.2}  
\begin{tabular}{ccccc}
\toprule
\multirow{2}{*}{\textbf{\makecell{Digit \\ Thickness}}} 
& \multicolumn{4}{c}{\textbf{Validation Accuracy (\%)}} \\ 
\cmidrule(lr){2-5}
& \textbf{Maj} & \textbf{Max} & \textbf{Top} & \textbf{Multi} \\
\midrule
1.4 & $80.2 \pm \text{\tiny 6.2}$ & $75.9 \pm \text{\tiny 1.4}$ & $71.3 \pm \text{\tiny 6.0}$ & $26.1 \pm \text{\tiny 13.4}$ \\
1.6 & $79.2 \pm \text{\tiny 3.7}$ & $75.7 \pm \text{\tiny 3.0}$ & $69.8 \pm \text{\tiny 3.8}$ & $21.2 \pm \text{\tiny 11.1}$ \\
1.8 & $58.4 \pm \text{\tiny 24.8}$ & $65.1 \pm \text{\tiny 18.0}$ & $54.1 \pm \text{\tiny 25.8}$ & $13.7 \pm \text{\tiny 11.6}$ \\
2.0 & $10.2 \pm \text{\tiny 1.1}$ & $26.5 \pm \text{\tiny 1.6}$ & $9.4 \pm \text{\tiny 0.9}$ & $0.0 \pm \text{\tiny 0.0}$ \\
2.2 & $10.9 \pm \text{\tiny 0.0}$ & $27.4 \pm \text{\tiny 0.0}$ & $10.1 \pm \text{\tiny 0.1}$ & $0.0 \pm \text{\tiny 0.0}$ \\
2.4 & $10.9 \pm \text{\tiny 0.1}$ & $27.4 \pm \text{\tiny 0.0}$ & $10.2 \pm \text{\tiny 0.6}$ & $0.0 \pm \text{\tiny 0.0}$ \\
2.6 & $15.7 \pm \text{\tiny 8.4}$ & $29.8 \pm \text{\tiny 11.2}$ & $13.0 \pm \text{\tiny 5.1}$ & $0.1 \pm \text{\tiny 0.1}$ \\
\bottomrule
\end{tabular}
\caption{Validation accuracy average of IPS over three random seeds, varying the thickness of the noise digit used in the Megapixel MNIST dataset. Experiments were run on 2000 training samples.}
\label{digitthick}
\end{table}

For task "Max", where the goal is to find the highest digit out of five digits with three digits being identical, we observed that the number 9 is predicted most often, resulting in a random accuracy of about 29\%. To find out why, we frame the task as finding the probability that a number is the maximum in a set of three different randomly chosen digits. The number of ways to choose 3 different numbers from 10 can be computed by the binomial coefficient $\binom{10}{3} = 120$. The number of possible ways to choose 2 numbers that are inferior to x is $\binom{x-1}{2}$. The probability that $x$ is the highest among the 3 chosen numbers is therefore $\binom{x-1}{2} / 120$. The probabilities from 0 to 9, are as follows \{0.00, 0.00, 0.01, 0.03, 0.05, 0.08, 0.13, 0.18, 0.23, 0.30\}. By predicting digit 9, the model maximizes its random performance which sits at 30\%.  

In all our experiments, the noise digit thickness was thus set to 1.925 to strike a balance between being able to converge while remaining as close to the threshold as possible.  

We further experimented with changing the number of noise digits on the canvas when their thickness is set to 1.925 (thickness used in all experiments). Results in Table \ref{validation_noise_levels} show that increasing the number of noise digits does not affect performance in cases with abundant data (5000 samples). For task "Maj" for instance when 100 noise digits are placed on the $3000 \times 3000$ canvas, validation sits at 99\% and remains consistent at 99\% with 800 noise digits on the canvas. This indicates that the main contributor of our proposed noise generation component to the performance is the thickness of the noise digit.

\begin{table}[htbp]
\centering
\setlength{\tabcolsep}{5pt}  
\renewcommand{\arraystretch}{1.2} 
\begin{tabular}{cccccc}
\toprule
\multirow{2}{*}{\textbf{Noise Amount}} 
& \multicolumn{4}{c}{\textbf{Validation Accuracy (\%)}} 
& \multirow{2}{*}{\textbf{Loss}} \\ 
\cmidrule(lr){2-5}
& \textbf{Maj} & \textbf{Max} & \textbf{Top} & \textbf{Multi} & \\
\midrule
100 & $99.0$ & $94.5$ & $92.7$ & $83.6$ & $0.195$ \\
200 & $98.8$ & $93.9$ & $93.6$ & $84.0$ & $0.193$ \\
300 & $99.0$ & $94.2$ & $94.4$ & $83.8$ & $0.194$ \\
400 & $98.9$ & $94.6$ & $92.5$ & $82.7$ & $0.199$ \\
600 & $99.0$ & $94.2$ & $94.2$ & $82.6$ & $0.194$ \\
800 & $99.1$ & $92.7$ & $93.5$ & $83.4$ & $0.211$ \\
\bottomrule
\end{tabular}
\caption{Validation accuracy of IPS trained on 5000 samples of a \(3000 \times 3000\) canvas with varying numbers of noise components. The sizes of the digits and noise components are both \(28 \times 28\).}
\label{validation_noise_levels}
\end{table}

\subsection{Effect of patch size}
\label{Patchsize}
To improve IPS's ability to generalize in low data settings with tiny object-to-image ratios, we experiment with varying the patch size to be higher or lower than the region of interest (ROI). The main finding of this section is that choosing the appropriate patch size becomes more important when less data is available. 

We find that a patch size that succeeds in solving the megapixel MNIST and Swedish traffic signs datasets with abundant data doesn't necessarily generalize to a low data scenario. By tuning the patch size we consistently achieved higher validation accuracy with IPS compared to the original patch sizes used by \cite{bergner_iterative_2023} to solve the full megapixel MNIST and Swedish traffic signs dataset. 

\subsubsection{Megapixel MNIST}
For the megapixel MNIST dataset, a canvas size of \(3000 \times 3000\) was used with a fixed digit size of \(84 \times 84\) and 1000 training samples. The patch sizes were changed to the following values: \(\{25\times25, 50\times50, 100\times100, 150\times150\}\) and the validation accuracy averaged over three random seeds. The range of patches were chosen to be larger and smaller than the size of the region of interest (ROI). Table \ref{patchsize} illustrates that the accuracy for all tasks is higher when the patch size is smaller than the ROI for the megapixel MNIST. When the patch size sits at 25, the validation accuracy for tasks "Maj", "Max", and "Top" is 76\%, 69\%, 72\%, while with a patch size of 150, the accuracy falls to 43\%, 47\%, and 40\% respectively. Interestingly, the highest validation accuracy for all tasks is observed when the patch size is set to \(25 \times 25\), except for the task "Multi" where a patch size of \(50 \times 50\) yields the highest testing accuracy (25\%).

\begin{table}[htbp]
\centering
\setlength{\tabcolsep}{1pt}  
\renewcommand{\arraystretch}{1} 
\begin{tabular}{cccccc}
\toprule
\multirow{2}{*}{\scriptsize \textbf{\makecell{Patch \\ Size}}} 
& \multirow{2}{*}{\scriptsize \textbf{\makecell{O2P \\ (\%)}}} 
& \multicolumn{4}{c}{\textbf{\scriptsize Validation Accuracy (\%)}} \\
\cmidrule(lr){3-6}
& & \textbf{\scriptsize Maj} & \textbf{\scriptsize Max} & \textbf{\scriptsize Top} & \textbf{\scriptsize Multi} \\
\midrule
25  & 1130 & $76.0 \pm \text{\tiny 9.7}$  & $69.3 \pm \text{\tiny 7.7}$  & $72.2 \pm \text{\tiny 18.4}$ & $14.2 \pm \text{\tiny 9.1}$ \\
50  & 282  & $63.6 \pm \text{\tiny 20.2}$ & $56.2 \pm \text{\tiny 13.0}$ & $58.7 \pm \text{\tiny 19.1}$ & $25.4 \pm \text{\tiny 14.4}$ \\
100 & 71  & $52.3 \pm \text{\tiny 21.0}$ & $55.4 \pm \text{\tiny 15.9}$ & $46.8 \pm \text{\tiny 18.5}$ & $7.6 \pm \text{\tiny 11.0}$ \\
150 & 31  & $43.0 \pm \text{\tiny 28.8}$ & $46.8 \pm \text{\tiny 22.4}$ & $40.2 \pm \text{\tiny 25.3}$ & $6.9 \pm \text{\tiny 11.0}$ \\
\bottomrule
\end{tabular}
\caption{Average validation accuracy with standard deviations of IPS for different patch sizes on the Megapixel MNIST. Each iteration was run for 100 epochs with an object-to-image ratio of 0.078\% using 1000 training samples on a \(3000 \times 3000\) canvas with a digit size of \(84 \times 84\).}
\label{patchsize}
\end{table}

In the megapixel MNIST implementation of IPS with 5000 training instances, the patch size was set to \(50 \times 50\) and the digit size to \(28 \times 28\), resulting in a \(31\%\) object-to-patch ratio (O2P). This is equivalent to our setup when the object size is \(84 \times 84\) and the patch size is \(150 \times 150\) as the object-to-patch ratio is \(84^2 / 150^2 = 31\%\). However, an object-to-patch ratio of \(1130\%\)  corresponding to a \(25 \times 25\) patch size (refer to Table \ref{patchsize}), is preferred over \(31\%\). The validation accuracy for an O2P ratio of \(1130\%\) (Patch size; \(25 \times 25\)) is 76\%, 69\%,  72\%, and 14\%  for tasks "Maj", "Max", "Top" and "Multi" and falls to 43\%, 47\%, 40\% and 7\%  respectively with an O2P ratio of \(31\%\) (Patch size; \(150 \times 150\)). An alternative way to set up the experiment is to keep the digit size the same as in IPS, however, this would have limited our ability to scale the patch size well below the original object size of \(28 \times 28\).

\subsubsection{Swedish traffic signs}
MNIST digits are all the same size across the dataset. To consider the sensitivity of a dataset with images containing different object-to-image ratios, we can look at the Swedish traffic signs dataset \citep{larsson_using_2011}, where the task is to classify traffic signs in road images. The original dataset contains 3,777 images with dimensions \(960 \times 1280\) and 20 different traffic sign classes. We employ the setup used in previous works \citep{katharopoulos_processing_2019, bergner_iterative_2023},  where the task is limited to detecting speed limits of 80, 70, and 50. This results in using only a subset of the original data, including 747 training and 684 validation instances. The subset includes 100 images for each speed limit class and 400 road images that do not contain a sign. 

After reproducing the results of IPS, we found that the dataset can be solved easily. To create a more challenging testbed, we experimented with 3 subset sizes across 5 model runs, namely \{25\%, 50\%, 100\%\} resulting in the following training set sizes \{744, 372, 184\}. Each instance is randomly sampled without replacement from the original set stratified by label: \emph{no sign} (50\%), \emph{80} (16.6\%), \emph{70} (16.6\%), \emph{50} (16.6\%). 

\begin{table}[h!]
\centering
\setlength{\tabcolsep}{3.5pt}  
\renewcommand{\arraystretch}{1.2}  
\begin{tabular}{cccccc}
\toprule
\multirow{2}{*}{\textbf{\scriptsize Subset}} 
& \multicolumn{2}{c}{\textbf{\scriptsize Validation}} 
& \multicolumn{2}{c}{\textbf{\scriptsize Training}} 
& \multirow{2}{*}{\scriptsize (s)/B} \\ 
\cmidrule(lr){2-3} \cmidrule(lr){4-5}
& \textbf{\scriptsize n} & \textbf{\scriptsize Mean (\%)} 
& \textbf{\scriptsize n} & \textbf{\scriptsize Mean (\%)} & \\
\midrule
25\%  & 169  & $79.745 \pm \text{\tiny 5.2}$ & 184  & $99.88 \pm \text{\tiny 0.2}$ & $3.25$ \\
50\%  & 341  & $95.223 \pm \text{\tiny 3.3}$ & 372  & $99.946 \pm \text{\tiny 0.1}$ & $7.15$ \\
100\% & 684  & $98.313 \pm \text{\tiny 0.2}$ & 747  & $99.917 \pm \text{\tiny 0.0}$ & $14.9$ \\
\bottomrule
\end{tabular}
\caption{Mean and standard deviation of IPS with different subsets of the Swedish traffic signs data, stratified to the following proportions: \emph{no sign} (50\%), \emph{80} (16.6\%), \emph{70} (16.6\%), and \emph{50} (16.6\%). (s)/B refers to seconds per batch.}
\label{train_val_stats}
\end{table}

Results in Table  \ref{train_val_stats} show that with  25\% of the data, the validation accuracy sits at 80\% as opposed to 95\% and 98\% for 50\% and 100\% of the training data. Additionally, the standard deviation for 25\% of the data sits at  5\% indicating that the model converged to a similar accuracy for each run. We therefore choose to train on 25\% of the data as it struggles with some harder instances in the validation set while consistently converging. This is in line with \cite{pawlowski_needles_2020}'s work who outlined the importance of finding low data solutions for small O2I problems.

We experimented on the following patch sizes: \(\{25 \times 25, 50 \times 50, 75 \times 75, 100 \times 100\}\), chosen as the range contains the generalization maxima. Each patch size was run for 5 random seeds, and the validation accuracy was averaged and reported in Table \ref{PASHM}. 

\begin{table}[h!]
\centering
\setlength{\tabcolsep}{1.1pt}  
\renewcommand{\arraystretch}{1.3} 
\begin{tabular}{cccc}
\toprule
\textbf{\scriptsize \makecell{Patch \\ Size}} 
& \textbf{\scriptsize \makecell{Validation \\ Accuracy (\%)}} 
& \textbf{\scriptsize \makecell{Time \\ (ms/Batch)}} 
& \textbf{\scriptsize \makecell{Peak Memory \\ (GB)}} \\
\midrule
25  & $71.12 \pm \text{\tiny 12.7}$ & $449.72$ & $0.97$ \\
50  & $77.75 \pm \text{\tiny 4.7}$  & $302.31$ & $1.11$ \\
75  & $84.62 \pm \text{\tiny 2.8}$  & $299.31$ & $1.33$ \\
100 & $79.29 \pm \text{\tiny 3.0}$  & $322.67$ & $1.65$ \\
\bottomrule
\end{tabular}
\caption{Mean and standard deviation of the validation accuracy of IPS on the Swedish traffic signs benchmark for different patch sizes. Each model was run 5 times for 150 epochs on a 25\% subset of the data. Time per epoch was taken by averaging each 16 batches in one epoch.}
\label{PASHM}
\end{table}

Generalization is highest when the patch size is 75 (85\%) (refer to Table \ref{PASHM}) and at its lowest when it is 25 (71\%) and 50 (78\%). For reference, the original setup of IPS set the patch size to \(100 \times 100\), yet within a low data setting with 25\% of the original size of the dataset, a patch size of 75 is preferred, with a marginal decrease in standard deviation. Specifically, for a patch size of \(100 \times 100\), the average validation accuracy is 79\% (Std: 3\%), while it climbs to 84\% for a patch size of \(75 \times 75\) (Std: 2.8\%). 

An additional benefit is the lower memory and runtime requirements of a smaller patch size. This is expected as the number of patches in memory \(M\) remains fixed in IPS. When the patch size decreases, fewer pixels are kept in memory. A patch size of \(100 \times 100\) consumes 1.6 GB of peak memory usage with an average time per batch of 323 ms. A patch size of \(75 \times 75\) on the other hand consumes 1.3 GB of peak memory with an average time per batch at a lower 299 ms. 

The implications for this are clear, in low data scenarios we can improve validation accuracy for low object-to-image ratio problems by tuning the patch size. By using the same object-to-image (O2I),  and object-to-patch (O2P) ratio we were able to get substantial gains in validation accuracy compared to the origin setup in IPS \citep{bergner_iterative_2023}. 

\subsection{Attention maps}
Here, we showcase attention maps for four instances in the megapixel MNIST dataset with different O2I ratios and 2 instances in the Swedish traffic signs data set with different O2I ratios and patch sizes. The purpose is to visualize the lack of robustness of IPS to low O2I ratios and the superiority of a smaller patch size relative to the ROI. 

\subsubsection{Megapixel MNIST}
\begin{figure}[htbp]
    \centering
    \begin{minipage}{\linewidth}
        \centering
        \includegraphics[width=\linewidth]{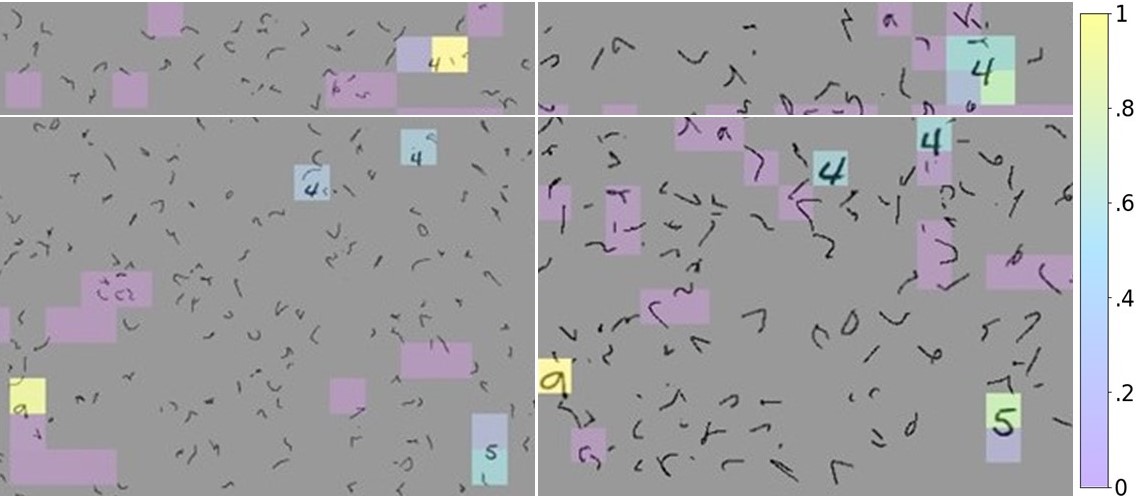} 
        \caption{Attention maps for different object-to-image ratios: 0.01\% (left) and 0.034\% (right) on a $1500 \times 1500$ canvas, with 800 noise digits on the left and 600 on the right. The maps display the top $M$ (100) most informative patches at the end of a full forward pass with IPS. The digit and noise size on the left is $28 \times 28$ and on the right, $56 \times 56$. Each image was truncated into two separate parts, one at $149 \times 645$ and the over at $482 \times 645$ to visualize all of the MNIST digits on the $1500 \times 1500$ canvas.}
        \label{map_1}
    \end{minipage}
\end{figure}

\begin{figure}[htbp]
    \centering
    \begin{minipage}{\linewidth}
        \centering
        \includegraphics[width=\linewidth]{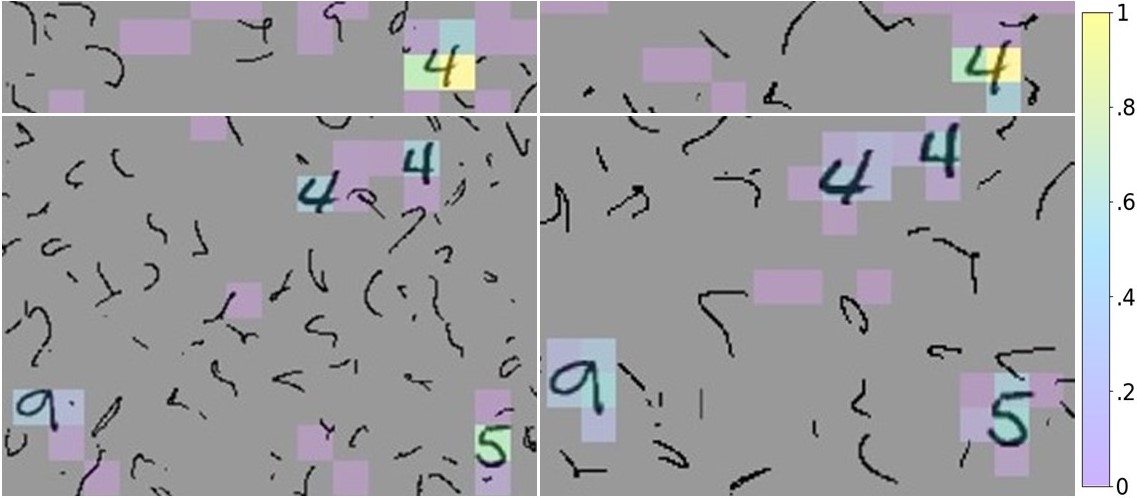} 
        \caption{ Attention maps for different object-to-image ratios: 0.078\% (left) and 0.13\% (right) on a $1500 \times 1500$ canvas, with 400 noise digits on the left and 200 on the right. The maps display the top $M$ (100) most informative patches at the end of a full forward pass with IPS. The digit and noise size on the left is $84 \times 84$ and on the right, $112 \times 112$. Each images was truncated into two separate parts, one at $149 \times 645$ and the over at $482 \times 645$ to visualize all of the MNIST digits on the canvas.}
        \label{map_2}
    \end{minipage}
\end{figure}

For the megapixel MNIST benchmark, four maps were produced by running identical models for 60 epochs, with 1000 training samples and the following object-to-image ratios: 0.01\% (left of Figure \ref{map_1}), 0.034\% (right of Figure \ref{map_1}), 0.078\% (left of Figure \ref{map_2}), and 0.13\% (right of Figure \ref{map_2}). 

The attention maps demonstrate that the lower the object-to-image ratio, the more uncertain the attention maps are in distinguishing between the noise and MNIST digits. All models successfully solve the task; however, the lowest object-to-image ratio (0.01\%) (left of Figure \ref{map_1}) includes many informative patches (in blue/purple with attention $<$ 0.6). These patches often overlay noise, as indicated by the number of colored patches highlighting noise digits. On the other hand, at the highest object-to-image ratio (0.13\%) in Figure \ref{map_2} on the right, there are no informative patches that cover noise.

This pattern is consistent with the O2I ratios in the center of both extremes, where at 0.034\% (right of Figure \ref{map_1}), more patches cover noise digits than at 0.078\% (left of Figure \ref{map_2}). The implications are that at higher object-to-image ratios, IPS is able to make clearer distinctions between the noise and the digits compared to lower O2I ratios where the difference is less pronounced.

\subsubsection{Swedish traffic signs}
\begin{figure}[htbp]
    \centering
    \begin{minipage}{\linewidth}
        \centering
        \includegraphics[width=\linewidth]{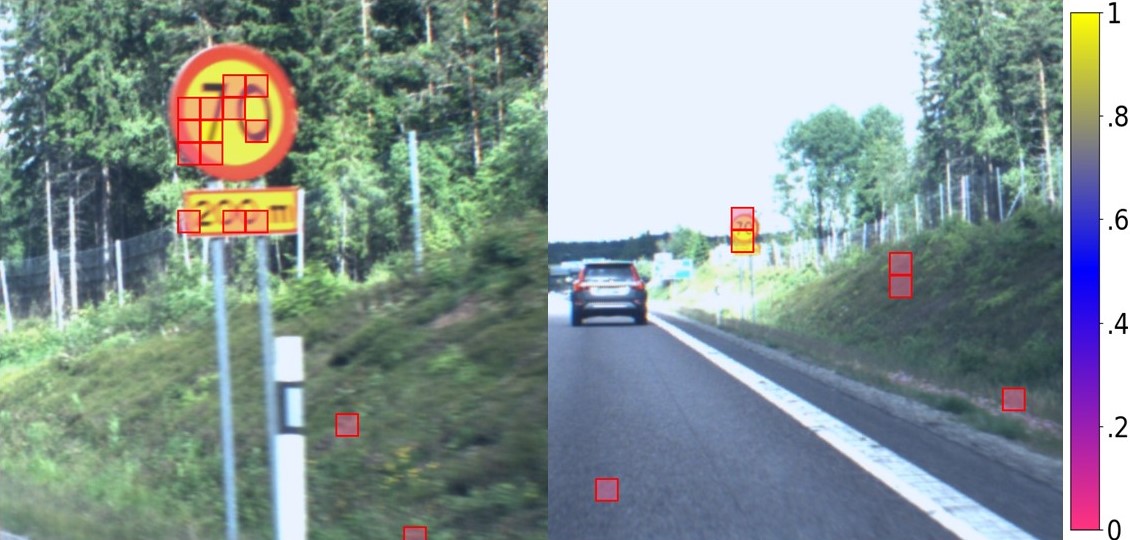} 
        \caption{Right: Attention map for image 86 (validation set) of the Swedish traffic signs dataset. Left: Attention map for image 82 (validation set). IPS was run for 140 epochs with a patch size and stride of 25. The memory buffer M is set to 20 patches. All patches M selected by IPS are outlined in red and the color corresponds to the normalized attention value. Each image was truncated to  $475 \times 480$  for ease of visualization and includes all patches with attention greater than 0.1. }
        \label{attentioj_map_1}
    \end{minipage}
\end{figure}

\begin{figure}[htbp]
    \centering
    \begin{minipage}{\linewidth}
        \centering
        \includegraphics[width=\linewidth]{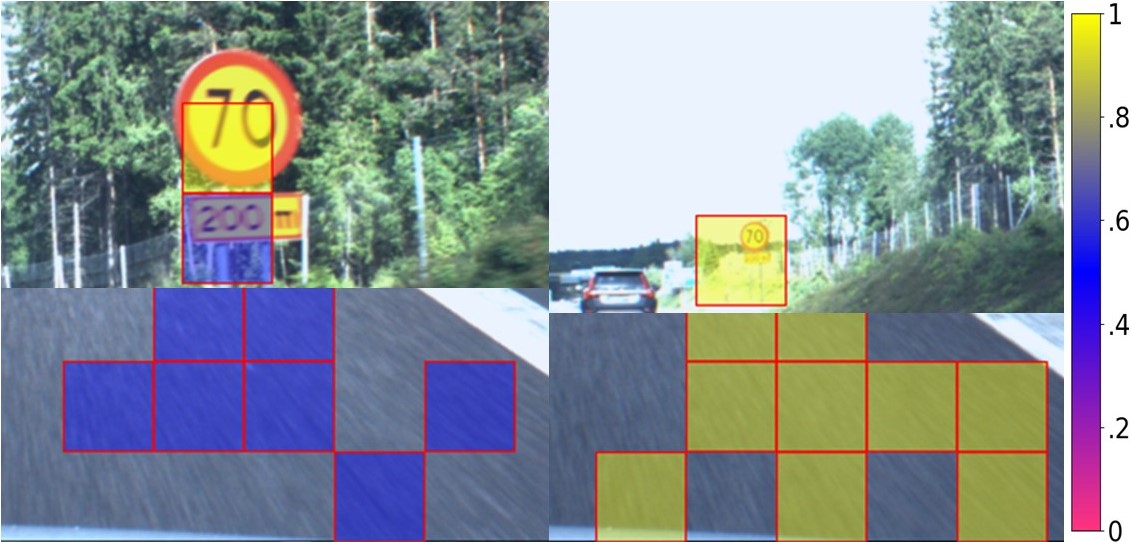} 
        \caption{Right: Attention map for image 86 (validation set) of the Swedish traffic signs dataset. Left: Attention map for image 82 (validation set). IPS was run for 140 epochs with a patch size and stride of 100. The memory buffer M is set to 20 patches. All patches M selected by IPS are outlined in red and the color corresponds to the normalized attention value. Each image was truncated into two parts to visualize all informative patches on the canvas with attention greater than 0.1. The leftmost images are of dimensions $232 \times 480$ and $243 \times 480$. The rightmost images are of dimension $241 \times 480$ and $234 \times 480$.}
        \label{attentioj_map_2}
    \end{minipage}
\end{figure}

In the Swedish traffic signs benchmark, the object-to-image ratio changes (from approximately 0.05\% to 0.4\%) as some signs are further away from the cameras than others. We trained two models: the first with a patch size and stride of $25 \times 25$, and the second with a patch size and stride of $100 \times 100$. The model weights were taken after 140 epochs of training, and the maps were plotted on images 86 and 82 of the Swedish traffic signs dataset. Both images contain the same scene but with different object-to-image ratios, 0.4\% for image 86 (right of Figures \ref{attentioj_map_1} and \ref{attentioj_map_2}) and 0.05\% for image 82 (left of Figure \ref{attentioj_map_1} and \ref{attentioj_map_2}). The attention values are normalized before plotting.

Figure \ref{attentioj_map_1} illustrates that when the patch size is lower than or equal to the smallest region of interest, inference is more stable across object-to-image ratios. When the patch size is fixed at $25 \times 25$, in Figure \ref{attentioj_map_1}, we can see that the model is able to find clusters of patches where a single one is more informative than the others. 

On the right of Figure \ref{attentioj_map_1} (O2I ratio: 0.05\%), there is a cluster of 2 patches overlaying the sign where one (in yellow) is significantly more informative than all patches in the image. In the left of Figure \ref{attentioj_map_1} (O2I ratio: 0.4\%), there are 12 patches covering the region of interest where a single patch (yellow) is more informative than all other patches.

In contrast, in Figure \ref{attentioj_map_2}, the patch size of $100 \times 100$ is substantially bigger than the smallest object. In the right of Figure \ref{attentioj_map_2} (O2I ratio: 0.05\%), the object of interest $(27 \times 27)$ occupies less than one-third of the patch $(100 \times 100)$. The image shows that all of the informative patches in the image are in the same range of attention (yellow) meaning IPS fails to find the single patch that solves the problem.  On the other hand, when the object-to-image ratio increases to 0.4\% (left of Figure \ref{attentioj_map_2}), a single patch is highlighted as informative.

This is undesirable as being able to delineate salient regions and non-informative patches is vital in certain domains such as digital histopathology. These observations further imply that a smaller patch size is advantageous in an inference setting as IPS can more easily delineate between informative and non-informative regions at low O2I ratios.

\subsection{Supplementary Findings }
\vspace{-0.5em}
As part of our preliminary experiments, we scaled the digit resolution both downward and upward from the original resolution of $28 \times 28$ with the following resolutions $\{14 \times 14, 21 \times 21, 28 \times 28, 46 \times 46, 64 \times 64\}$ with a canvas size of $1500 \times 1500$.
 We ran these configurations for 150 epochs, resulting in the results presented in Table \ref{digit_res_bars_2}.

\begin{table}[h!]
\centering
\setlength{\tabcolsep}{6pt} 
\renewcommand{\arraystretch}{1.2}  
\resizebox{\columnwidth}{!}{  
\begin{tabular}{ccccc}
\toprule
\multirow{2}{*}{\textbf{\scriptsize \makecell{Digit \\ Resolution}}} 
& \multirow{2}{*}{\textbf{\scriptsize \makecell{O2I \\ (\%)} }} 
& \multicolumn{3}{c}{\textbf{\scriptsize Validation Accuracy (\%)}} \\ 
\cmidrule(lr){3-5}
& & \textbf{\scriptsize Maj} & \textbf{\scriptsize Max} & \textbf{\scriptsize Top} \\
\midrule
14 x 14 & 0.008 & $66.5$ & $60.1$ & $49.5$ \\
21 x 21 & 0.012 & $74.7$ & $78.4$ & $58.9$ \\
28 x 28 & 0.035 & $99.1$ & $93.7$ & $92.7$ \\
46 x 46 & 0.094 & $98.9$ & $93.3$ & $92.2$ \\
64 x 64 & 0.182 & $99.3$ & $92.9$ & $94.6$ \\
\bottomrule
\end{tabular}}
\caption{Validation accuracy of IPS using lower resolution digits to vary the object-to-image (O2I) ratio. The model was trained for 150 epochs on a \(1500 \times 1500\) canvas with 5000 training samples and the corresponding object-to-image ratios.}
\label{digit_res_bars_2}
\end{table}

\begin{figure}[htbp]
    \centering
    \includegraphics[width=\columnwidth]{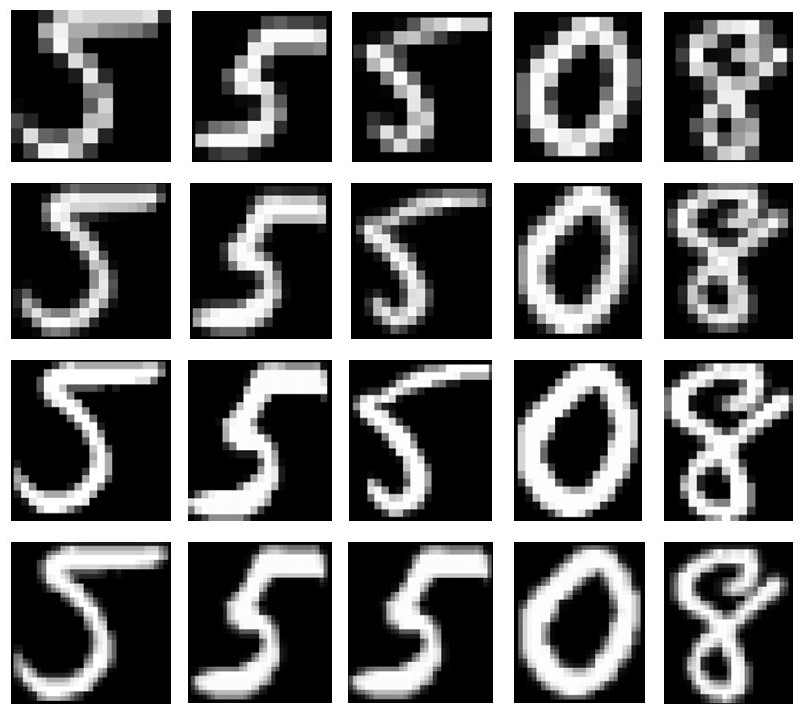} 
    \caption{Five MNIST digits (left to right) from the Megapixel MNIST dataset, downsampled to the following resolutions in order from top to bottom: \(\{14 \times 14, 21 \times 21, 28 \times 28, 46 \times 46, 64 \times 64\}\).}
    \label{digit_res}
\end{figure}

 We can see that there is a stark decrease in performance when the digit size is down-sampled below its original size $\{28 \times 28\}$. When the digit size is set to $14 \times 14$ (O2I: 0.008\%), the validation performance for the task "Maj" sits at $66\%$ for $5000$ samples and $150$ epochs. In contrast, the same O2I of $0.01\%$ without scaling the digit size down, with a $28 \times 28$ digit on a $3000 \times 3000$ canvas, needs $1000$ training samples to achieve a validation score of $66\%$. Unsurprisingly, when the image is downsampled below its original size, the loss of information requires more data to achieve comparable validation performance (refer to Figure \ref{digit_res} for a visual representation of the down-sampled digits). This observation motivates our choice to scale the image to $3000 \times 3000$ and ensure that we only scale the digits and noise upwards.

\section{Discussion \& Conclusion}
Our work introduces an extension of the Megapixel MNIST benchmark that includes a novel method for increasing and decreasing the object-to-image ratio as well as a more adverse noise generation strategy. Experimental results generalize the observations made on CNNs to IPS whereby the O2I threshold below which the classifier fails to generalize is affected by the training dataset size. We further observe that the magnitude of this interaction differs for each task of the Megapixel MNIST.  For tasks "Maj" and "Top", the rate is at its highest, followed by tasks "Max" and "Multi" where in the latter, this rate is almost at 0. 

Our findings relating to the patch size indicate that in a low-data setting, tuning the patch size results in an average improvement of validation performance of 15\% for the megapixel MNIST and + 5\% for the Swedish traffic signs dataset  compared to the original object-to-patch ratios in IPS. In both datasets, a patch size that is smaller than the original implementation is preferred. For the Swedish traffic signs, we also empirically demonstrate that the smaller patch size results in less memory consumption (1.6 GB of peak memory for the original patch size of 100 and 1.3 GB for our tuned patch size of 75). Lastly,  our findings confirm the suspicions of \cite{pawlowski_needles_2020}. At lower object-to-image ratios, the model struggles to converge with noise that more closely resembles the distribution of the region of interest.  

 We also experimented with two regularization methods, STKIM and Diversity regularization \citep{zhang_attention-challenging_2024}, which despite extensive tuning, resulted in no improvement of validation accuracy.  Future works should consider other memory-efficient patch-based classifiers such as Attention sampling \citep{katharopoulos_processing_2019} and Differentiable Top-K \citep{cordonnier_differentiable_2021} to use as baselines.  

\section*{Declaration}

\subsection*{Data and code availability }
The source code as well as the results for each model run in this work has been made fully available at \url{https://github.com/MRiffiAslett/ips_MaxRiffiAslett.git}. 

\subsection*{Acknowledgments}
We extend our gratitude to Dr. Stuart Norcross for his help with the GPU setup and connection to the university server.

\bibliography{sn-bibliography}
\end{document}